# Creating a large language model of a philosopher

Eric Schwitzgebel[1], David Schwitzgebel[2], and Anna Strasser[3,4]

[1]Department of Philosophy, University of California, Riverside
[2]Institut Jean Nicod, École Normale Supérieure, Université PSL
[3]Faculty of Philosophy, Ludwig-Maximilians-Universität München
[4]DenkWerkstatt Berlin

**Correspondence**

Eric Schwitzgebel, Department of Philosophy, University of California, Riverside, CA 92521, United States.

Email: eschwitz@ucr.edu

**Funding information**

University of California, Riverside, Academic Senate Grant

Can large language models produce expert-quality philosophical texts? To investigate this, we fine-tuned GPT-3 with the works of philosopher Daniel Dennett. To evaluate the model, we asked the real Dennett ten philosophical questions and then posed the same questions to the language model, collecting four responses for each question without cherry-picking. Experts on Dennett's work succeeded at distinguishing the Dennett-generated and machine-generated answers above chance but substantially short of our expectations. Philosophy blog readers performed similarly to the experts, while ordinary research participants were near chance distinguishing GPT-3's responses from those of an "actual human philosopher."

# 1. INTRODUCTION

Artificial Intelligence can now outperform even expert humans in games such as chess, go, and poker and in practical domains such as lung cancer screening, predicting protein structure, and discovering novel matrix multiplication algorithms (Campbell, 2002; Silver et al., 2016, 2018; Ardila et al., 2019; Brown & Sandholm, 2019; Jumper et al., 2021; Fawzi et al., 2022). ChatGPT has received considerable popular attention for its capacity to generate passable short student essays (Huang, 2023). But presumably expert-level professional philosophy requires a type of expertise that renders it safe from AI takeover—at least for a while. Machines will not soon, it seems, generate essays that survive the refereeing process at *Philosophical Review*. We sought to explore *how* safe philosophy is. How close can we get to developing an AI that can produce novel and seemingly intelligent philosophical texts? The question is of interest for what it reveals about the nature of both mind and language.

    Concerning language, it might be thought that the capacity to generate novel, interpretable, and structurally sophisticated prose—unless that prose is the mere (stochastic) parroting of pre-existing material (Bender et al., 2021)—would require symbolically encoded, innate linguistic capacities (e.g., Chomsky, 2007; Berwick et al., 2011); or at least that it would require "functional" competence via dedicated mechanisms for formal reasoning, pragmatics, knowledge of interlocutors, and so forth, which the best known and most successful AI systems generally lack (Mahowald et al., 2023; Shanahan, 2023). Concerning mind, it might be thought that the capacity to generate novel and seemingly insightful seeming-philosophy, concerning for example the nature of consciousness, is particularly telling evidence of humanlike consciousness and cognitive sophistication (e.g., Descartes, 1649/1991; Davidson, 1984; Schneider, 2019). Furthermore, to the extent ordinary technology users become unable or unwilling to distinguish artificial systems from human systems, they might similarly become unable or unwilling to attribute different mental properties to the two types of systems. The most famous example of a discriminability criterion is, of course, the Turing Test (originally proposed by Alan Turing in 1950 as an "imitation game"), where attribution of "thought" (Turing, 1950) or "consciousness" (Harnad, 2003) is purportedly justified when the artificial system's responses are indistinguishable from the responses of a human.

The past few years have seen swift progress toward sophisticated, human-like language generation. Natural language processing is a booming subfield of AI research, with notable successes in automatic translation (DeepL), computer code generation (GitHub Copilot), lipreading (LipNet; Assael et al., 2016), and producing original prose with fluency similar to that of a human (Steven & Iziev, 2022). In June 2022, Google's LaMDA model made international news when Google engineer Blake Lemoine said he became convinced that LaMDA was sentient after engaging in philosophical dialogue with it (Hofstadter, 2022; Klein, 2022; Roberts, 2022; Tiku, 2022). Another recent model, ChatGPT, has rapidly risen into prominence as a remarkably fluent chat-bot, capable of producing prose that can pass as, for example, excellent student essays, raising concerns about plagiarism and cheating (Herman, 2022; Hutson, 2022; Marche, 2022; Huang, 2023).

Arguably, chess moves and formulaic summaries of the themes of Hamlet are one thing, and creative philosophical thinking is quite another. The fact that it is sometimes difficult to distinguish machine-generated text from human-generated text is not new: In social media, electronic customer service, and advertising, we are increasingly confronted with machine-produced content that is easily mistaken for human-produced content. Empirical research sometimes makes use of this indistinguishability when experimental protocols with artificial agents are used to test hypotheses about humans' social cognitive mechanisms (Strasser, 2022). However, indistinguishability tests mean little for brief and unsophisticated outputs. Longer outputs are potentially more interesting. If existing language models can be shown in a rigorous, scientific study to approach professional human philosophers in language performance, it would force a theoretical choice upon those who see such sophisticated linguistic performance as indicative of genuine language capacities and/or genuine mentality: Either deny that the outputs, despite their apparent novelty and sophistication, are genuinely linguistic and reveal genuine mentality, or accept that large language models are capable of genuine language and genuine, even sophisticated, thought.

We aimed not to create a language model fully indistinguishable from a human or one that could generate publishable philosophical articles, but rather to take a small step in that direction by creating a language model that can produce paragraph-long texts that even philosophical experts would find difficult to distinguish from texts produced by a professional philosopher. We succeeded beyond our expectations, as we will detail in this article. In short, we created a language model of Daniel Dennett sufficiently convincing that experts in Dennett's work frequently mistook paragraphs generated by the language model for paragraphs actually written by Dennett.

Our project employed OpenAI's GPT-3, a 96-layer, 175-billion parameter language model trained on hundreds of billions of words of text from Common Crawl, WebText, books, and Wikipedia (Brown et al., 2020)— a large language model (LLM).[1] After it has been trained, GPT-3 uses textual input (prompts) to predict likely next "tokens"—sequences of characters that often co-occur in written text (the training data). Using these predictions, GPT-3 can generate long strings of text by outputting a predicted string, then using that output as part of the context to generate the next textual output. You can engage in seemingly intelligent conversations with it: If you ask a question, it will often (not always) generate a sensible-seeming answer. Notably, GPT-3 has responded with seemingly intelligent replies to philosophical discussions about artificial intelligence and consciousness (Wiseman, 2020; Zimmerman, ed., 2020; Schwitzgebel, 2021a), though it is likely that such impressive outputs typically involve a certain amount of "cherry-picking"—that is, having GPT-3 produce multiple outputs and then having a human pick the best among them. We aimed to evaluate the output of an LLM without cherry-picking.

In reviewing recent attempts to simulate human-generated text online using LLMs (e.g., Mahdawi & GPT-3, 2020; Araoz, 2021; Clarke, 2022), we have found that it is often difficult to precisely identify the underlying technology, the training methods, or how much cherry-picking was done. Even models with relatively well-documented architecture and pretraining processes generally rely on some human intervention and curation and are often best characterized as hybrid models, in contrast to "pure" autoregressive transformers such as GPT-3.

For example, an art project titled "Chomsky Vs Chomsky" presents a virtual version of Noam Chomsky—a location-based, mixed reality (MR) experience (Rodriguez, 2022) drawing on the publicly available texts and recorded lectures of Chomsky. This project offers the experience of asking questions orally in virtual reality and getting an audio answer almost indistinguishable from recordings of the real Chomsky. However, Chomsky Vs Chomsky also includes a set of prescribed complete responses triggered by certain keywords, and its outputs are highly curated, relying on hybrid techniques, explicit commands, cherry-picking, and other supplementations.

---

[1] Considering the current rapid development of LLMs, it must be pointed out here that GPT-3, in contrast to its various successors such as ChatGPT or LaMDA, is based solely on a transformer technique. This means that the ability to produce text is exclusively based on complex statistical evaluations of the training data. In contrast, LaMDA does not only generate potential responses, but all outputs are in addition filtered for safety, grounded on an external knowledge source, and re-ranked to find the highest-quality response (Thoppilan et al., 2022). More recent LLMs approach more and more the status of hybrid systems, which are based on neural networks but also contain parts of symbolic AI (Hilario, 1995). From a philosophical perspective, the LLMs solely based on a statistical self-attention mechanism are of particular interest, as they can show how far one can get without adding further abilities such as memory, human-evaluated quality estimations and the like.

Given the importance and visibility of LLMs like ChatGPT, it is surprisingly rare to see rigorous scientific testing of people's capacity to distinguish LLM outputs from human outputs. While there are many "automated" or "static" benchmarks that are used to assess the performance of artificial language models (e.g., Michael et al., 2022), these benchmarks do not reliably correspond to human judgments about the quality (or "human-ness") of AI-generated texts (van der Lee et al., 2019). Even the relatively few studies that do require humans to directly assess the quality of AI-generated text often are severely limited by small sample sizes, inter-rater disagreement, and inconsistent measurement scales (Novikova et al., 2017; van der Lee et al., 2019). Furthermore, these studies overwhelmingly focus on evaluating the quality of explicitly-labeled AI-generated content rather than directly assessing *human-machine discrimination*—the ability of humans to accurately distinguish, without explicit labeling, between machine-generated and human-generated content. To the extent that truly competent language models should be able to faithfully imitate human-generated text, this type of evaluation should provide a reliable and implicit measure of the quality of AI-generated text.

As of this writing, we are aware of only one study evaluating people's ability to distinguish recent LLM outputs from human-generated outputs, using rigorous psychological methods explained in detail: Clark et al. (2021). The authors collected human-composed short texts in three different domains: stories, news articles, and recipes. They then used the base models of GPT-3 and GPT-2 (an older, smaller model) to generate new texts within the same domains. The authors truncated and formatted the AI-generated texts to be comparable in length and format with the human-generated ones, but otherwise left them unedited. They then recruited 130 participants per domain and model, asking them to evaluate which texts were human-generated, providing an explanation for their judgment. Participants' accuracy in distinguishing GPT-3-generated from human-generated text was only 50%—not significantly different from chance. Follow-up experiments explored the qualitative explanations provided by participants and—based on these judgments—attempted to train participants to identify AI-generated texts. However, even the most effective training method did not bring accuracy above 57% in any domain. These results speak to the remarkable capacity of GPT-3 to generate texts that resemble human linguistic content. Brown et al. (2020) used a similar method, described more briefly and focused just on news articles, finding similar results, with moderately good discrimination rates for smaller language models and near-chance performance with the largest version of GPT-3.

Two other studies warrant brief mention. Gao et al. (2022) used ChatGPT to generate scientific abstracts, then recruited scientists to distinguish those abstracts from genuine scientific abstracts. Scientists were substantially above the 50% chance rate in discriminating machine-generated from human-generated abstracts, with at 32% false negative rate (classifying machine-generated texts as human-generated) and a 14% false positive rate (classifying human-generated texts as machine-generated). However, this study is limited by lack of transparency regarding the amount of cherry-picking or editing of the outputs; the employment of only four experts for the discrimination task, all of whom were authors of the article (thus possibly motivated differently than independent experts); lack of comparison between novice performance and expert performance; and non-parallelism between the machine and human language-generation tasks, which were not in response to the same inputs. Furthermore, ChatGPT is not a pure transformer architecture and thus operates on somewhat different underlying principles than GPT-3.

Dugan et al. (2020) presented online participants with sequences of sentences. The first sentences were human written, continuing with new sentences and eventually transitioning to machine-written sentences. Participants were instructed to discriminate at what point the text transitioned from human-written to machine-written. Only 16% of participants identified the exact boundary, and the average estimate was 1.9 sentences after the boundary, suggesting that machine-generated text can fool non-experts, but normally only briefly. However, raters were not experts, nor were they asked to discriminate human-generated from machine-generated texts in a side-by-side comparison. Furthermore, participant quality might have been low, with no financial motivation for correct responding and 25% of respondents excluded for failing a simple attention check.

While the base model of GPT-3 is impressive, it is not specialized to produce philosophical text. The studies by Clark et al. (2021) and Brown et al. (2020) were intended to assess the capabilities of the base model at some common tasks (stories, news articles, and recipes). It is not clear whether these results can be generalized to professional-quality philosophy, which is arguably less formulaic than recipes and short, generic news stories. Furthermore, the researchers relied on a non-expert sample of participants. Whatever the domain under evaluation, non-experts might be far worse than experts at distinguishing human-based output from the output of an LLM. We sought to see how close GPT-3 is to a much higher level of achievement, specifically in philosophy, by examining whether philosophical experts could distinguish the outputs of GPT-3 from the outputs of one of the world's best-known living philosophers.

In order to produce more specialized text, GPT-3 can be "fine-tuned" with custom training data. That is, it can be given additional training on a specific corpus so that its outputs reflect a compromise between GPT-3's default weightings and weightings reflecting the structure of the new corpus. Since non-fine-tuned GPT-3 can sometimes produce seemingly philosophical replies to textual inputs, we conjectured that a version of GPT-3 fine-tuned on the work of a particular philosopher might be able to speak in something like that philosopher's voice, seeming to express views consistent with the views of the philosopher on which it has been trained.

In the pilot phase of our project, we created several fine-tuned models, one based on the English translation of the works of Kant and another based on the corpus of a well-known philosophical blog (Eric Schwitzgebel's blog, The Splintered Mind, which has been running since 2006, with over a million words of philosophical content; Schwitzgebel, 2021b). For that piloting, we used two different GPT-3 models: Curie and Davinci. Both models are from the same general GPT-3 model family, but the Curie model is smaller, faster, and less powerful, while the Davinci model was the most powerful model then on offer by OpenAI. We observed impressive improvement between Curie and Davinci, but even the Curie model was able to produce outputs with substantial structure on a single philosophical theme, sometimes carrying unified, complex philosophical threads of argument in an organized structure over the course of several hundred words.

Finally, we fine-tuned the full Davinci model on most of the collected works of philosopher Daniel Dennett. Since it is difficult to measure the philosophical quality of outputs produced in this way, we employed a discrimination task. Thus, we established an indirect quality measure, which assumes that the distinguishability between machine-generated text and text generated by one of the world's best-known philosophers can give an indication of the quality of the machine-generated text. To investigate how easily the outputs of the fine-tuned GPT-3 could be distinguished from Dennett's real answers, we asked Dennett ten philosophical questions and then posed those same questions to our fine-tuned version of GPT-3 ("DigiDan"). Then we recruited experts in Dennett's work, blog readers, and ordinary online research participants into an experiment in which they attempted to distinguish Dennett's real answers from the answers generated by DigiDan. Participants also rated all answers, both Dennett's and DigiDan's, for similarity to "what Dennett might say" or "what a real human philosopher might say."

## 2. LANGUAGE MODEL OF DENNETT: DESIGN

### 2.1 Fine-tuning

For the purposes of this project, Dennett provided us with the entire digitally available corpus of his philosophical work. We converted most of this corpus (15 books and 269 articles) into segments of 2000 or fewer "tokens" for use as training data. (A token is a sequence of commonly co-occurring characters, with approximately ¾ of a word per token on average.) This process involved converting .pdf and word processing files into plain text format, stripping away headers, footnotes, scanning errors, marginalia, and other distractions, resulting in approximately three million tokens in 1,828 segments, including 254 segments from published interviews. On 11 March 2022, we fine-tuned the full GPT-3 Davinci engine on this corpus, using blank prompts and the 1,828 segments as completions, repeating the process four times (four epochs).

### 2.2 Prompt engineering

GPT-3 completions are highly sensitive to the content and structure of the prompts, and good "prompt engineering" is important for coaxing useful replies from GPT-3. After some exploratory testing, including several long and detailed prompts, we settled on the following simple prompt:

> Interviewer: [text of question]
> Dennett:

This simple prompt has several advantages: First, its minimal structure reduces potential concerns about the prompt possibly nudging completions toward specific philosophical content, as a more substantive prompt might. Second, it encourages the model to speak in the first person, voicing Dennett's views, rather than speaking in the third person about Dennett (possibly critically). Third, its simple format makes it easily generalizable to other cases.

**2.3 Question design**

We then designed ten questions addressing various themes across Dennett's corpus, including, for example, consciousness, free will, and God. The questions were somewhat complicated, and most contained more than one part, so as to invite complex answers from both Dennett and DigiDan. For example:

> What is a "self"? How do human beings come to think of themselves as having selves?

Before we produced the machine-generated answers, Dennett provided us with sincere written answers to all ten questions, ranging in length from 40 to 122 words.

**2.4 Collecting DigiDan's responses**

We collected the model's responses on the OpenAI playground. Before testing with our specific ten questions, we explored a variety of playground parameter settings—such as increasing or decreasing the "temperature" (the chance of lower-probability completions)— but we found no combination of settings that performed notably better than the default parameters (temperature = 0.7, top P = 1, frequency penalty = 0, presence penalty = 0, and best of = 1). Using the prompt described in Section 2.2, we then collected four responses from DigiDan for each of the ten questions.

We aimed to collect responses about the same length as Dennett's own responses to the same questions. Thus, if Dennett's response to a question was N words long, we excluded responses that were less than N-5 words long, counting a response as having ended either when the model reached a stop sequence or when it output "Interviewer," indicating the end of "Dennett's" answer and the beginning of the hypothetical interviewer's follow-up question. Two answers were excluded other than on grounds of length: one for describing Dennett's view in the third person and one for being potentially offensive. For eight of the ten prompts, zero to two outputs were excluded. However, for two of Dennett's longer answers, it took more than six attempts to generate four usable answers (16 total attempts in one case and 22 in another). The full list of questions and responses is available in the online supplement at https://osf.io/vu3jk.

Importantly, we never used perceived quality of response as a basis for selection. There was no "cherry-picking" of responses that we judged to be better, more Dennett-like, or more likely to fool participants.

**2.5 Editing DigiDan's responses**

To prevent guessing based on superficial cues, we replaced all curvy quotes with straight quotes, all single quotes with double quotes, and all dashes with standard em-dashes. We also truncated responses at the first full stop after the response achieved the target length of N-5 words. Apart from this mechanical editing, there was no editing of the model's responses.

**2.6 Research participants**

We recruited three groups of research participants. First, 25 *Dennett experts* were nominated by and directly contacted by Daniel Dennett or Anna Strasser. Second, 100 *ordinary research participants* were recruited for a payment of $3.00 each from Prolific Academic, a widely used source of psychological research participants, limited to US and UK participants with at least 100 Prolific completions, at least a 95% approval rate, and at least a bachelor's degree. Third, 304 *blog readers* were recruited from The Splintered Mind, with links from Twitter and Facebook, with no payment or required inclusion criteria. Two ordinary research participants were excluded for completing in fewer than four minutes, and two blog readers were excluded for completing in fewer than eight minutes, leaving 98 and 302 participants for analysis, respectively. One Dennett expert completed the survey twice, so only their first set of responses was included.

**2.7 Test structure: Experts' and blog readers' version**

Dennett experts and blog readers saw identical versions of the test (stimulus materials available at https://osf.io/vu3jk). After consenting, they were instructed as follows:

> In the course of this experiment, please do not consult any outside sources to help you answer the questions. Don't look things up on the internet. Don't look at books or notes you have. Don't consult with friends. Just do your best with what you already know.

Thereafter followed ten questions in two parts. Each question began as follows:

> We posed the question below to Daniel C. Dennett and also to a computer program that we trained on samples of Dennett's works. One of the answers below is the actual answer given by Dennett. The other four answers were generated by the computer program. We'd like you to guess: which one of the answers was given by Dennett?
> Question:

After the colon, we inserted the text of one question we had posed to Dennett and to our fine-tuned version of GPT-3. The order of the questions was randomized. After each question, five possible answers were presented, one by Dennett and four by DigiDan, in random order, and participants were instructed to guess which answer was Dennett's.

The second part of each task presented the question and all five answers again. Participants were instructed to rate each answer (Dennett's plus the four from DigiDan) on the following five-point scale:

> "not at all like what Dennett might say" (1)
> "a little like what Dennett might say" (2)
> "somewhat like what Dennett might say" (3)
> "a lot like what Dennett might say" (4)
> "exactly like what Dennett might say" (5)

The test concluded by asking highest level of education, country of residence, and "How much of Daniel C. Dennett's work have you read?" (response options: "I have not read any of Dennett's work," "I have read between 1 and 100 pages of Dennett's work," "I have read between 101 and 1000 pages of Dennett's work," "I have read more than 1000 pages of Dennett's work"). All questions were mandatory, so there were no missing data.

## 2.8 Test structure: Ordinary research participants' version

Ordinary research participants were assumed not to be familiar with Dennett's work, so the instructions referred instead to "a well-known philosopher" and participants were instructed

"select the answer that you think was given by the actual human philosopher." In the rating sections, "Dennett" was replaced with "a real human philosopher." The test concluded with questions about education level, country of residence, number of philosophy classes taken, and familiarity with the philosopher Daniel C. Dennett.

Since we thought ordinary research participants might find the task tiring, each was asked only five questions, randomly selected from the full set of ten. As an incentive to careful responding, participants were offered a $1 bonus if they guessed at least three of the five questions correctly.

**2.9 Hypotheses**

We hypothesized:

(1.) that expert respondents would perform better than ordinary research participants,
(2.) that expert respondents would on average guess correctly at least 80% of the time, and
(3.) that expert respondents would rate Dennett's actual answers as more Dennett-like than GPT-3's answers.

**2.10 Analytic method**

All inferential statistical analyses were conducted independently in R (version ID: 4.1.1; IDE: RStudio) and SPSS (Version 27.0.0.0), and the results were cross-verified between the two sources. All analyses were two-tailed, and $\alpha$ was set to .05. All one-sample $t$-tests are indicated with "$t()$," and all paired-samples $t$-tests are indicated with "paired $t()$." Two additional analyses were conducted: a one-proportion $z$-test (section 5) and an independent-samples $t$-test (section 4.3).

**3. LANGUAGE MODEL OF DENNETT: GENERAL RESULTS**

**3.1. Ordinary research participants**

The majority of ordinary research participants (58%) reported a bachelor's degree as their highest level of education, but a substantial minority (39%) reported an advanced degree. The majority (67%) reported having taken no philosophy classes, and only a few (5%) reported any graduate-level coursework in philosophy. A large majority (83%) reported not having heard of Daniel Dennett, and very few (4%) reported having read any of his work.

Overall, ordinary research participants responded correctly an average of 1.20 times out of 5, near the chance rate of 20%. A one-sample, two-tailed $t$-test did not identify a significant difference between participant scores and the chance guess rate of 20% (M = 1.20, $t(97)$ = 1.71, $p$ = .09, $d$ = .17, SD = 1.18, CI = [.97, 1.44]). Only 14% of the participants earned the bonus payment for guessing at least three correctly, and none guessed all five correctly.

On average, ordinary research participants rated both Dennett's actual answers and the model's answers as "somewhat like what a real human philosopher would say," with no statistically detectable difference in the ratings ($M_{Dennett}$ = 3.11, $M_{GPT-3}$ = 3.08, paired $t(97)$ = .47, $p$ = .64, $d$ = .05, $SD_{difference}$ = .69, $CI_{difference}$ = [-.10, .17])

Thus, ordinary research participants distinguished our fine-tuned GPT-3's answers from those of Dennett at rates at or near chance. For the most part, they were unable to distinguish DigiDan from an actual human philosopher.

## 3.2. Blog readers

The majority of blog reader respondents (57%) reported advanced degrees in philosophy, with 45% reporting PhDs. Only 6% reported not having read any of Dennett's work. The majority (64%) reported having read more than 100 pages of Dennett's work, and 18% reported having read over 1000 pages of Dennett's work.

Overall, blog readers responded correctly an average of 4.81 times out of 10 (48%), substantially above the chance rate of 20% (M = 4.81, $t(301)$ = 23.22, $p < .001$, $d$ = 1.34, SD = 2.10, CI = [4.57, 5.05]). They also rated Dennett's actual answers as significantly more Dennett-like than the model's answers ($M_{Dennett}$ = 3.60, $M_{GPT-3}$ = 2.65, paired $t(301)$ = 23.00, $p < .001$, $d$ = 1.32, $SD_{difference}$ = .72, $CI_{difference}$ = [.87, 1.03]).

Thus, blog readers—the majority of whom had graduate degrees in philosophy and substantial familiarity with Dennett's work—were able to distinguish Dennett's answers from those of our fine-tuned version of GPT-3 at rates substantially above chance, getting about half correct when given a five-alternative forced choice.

## 3.3. Dennett experts

The target group of greatest interest was the Dennett experts, most of whom (68%) reported having read over a thousand pages of Dennett's work. Overall, this group responded correctly an average of 5.08 times out of 10 (51%), significantly better than chance (M = 5.08, $t(24)$ = 7.13, $p < .001$, $d$ = 1.43, SD = 2.16, CI = [4.19, 5.97]). They also rated Dennett's actual answers as significantly more Dennett-like than the model's answers ($M_{Dennett}$ = 3.73, $M_{GPT-3}$ = 2.34, paired $t(24)$ = 8.44, $p < .001$, $d$ = 1.69, $SD_{difference}$ = .83, $CI_{difference}$ = [1.06, 1.74]) .

As these numbers suggest, the Dennett experts did not detectably outperform the blog readers ($M_{experts}$ = 5.08, $M_{blog}$ = 4.81, $t(325)$ = .62, $p$ = .54, $d$ = .13, SD = 2.11, CI = [-.59, 1.13]). Although experts were able to distinguish Dennett's answers from DigiDan's at rates significantly better than chance, like our blog readers, they only got about half correct for this five-alternative forced-choice task (Figure 1). This value is significantly below the hypothesized accuracy of 80% (M = 5.08, $t(24)$ = -6.76, $p < .001$, $d$ = -1.35, SD = 2.16, CI = [4.19, 5.97]), contradicting our initial hypothesis.

INSERT FIGURE 1 HERE

## 4. LANGUAGE MODEL OF DENNETT: ANALYSIS OF SELECTED QUESTIONS

Not all of the questions were equally difficult. Although ordinary research participants scored at about chance on average, their responses differed from chance on some individual questions. For example, on the question about whether robots can have beliefs (Robot), 38% guessed correctly (18/47, one-proportion $z$ = 3.14, $p$ = .002).

Of greater interest, the performance of experts also varied by question. The two questions with the highest rate of expert success—80% and 76%, respectively—were the Chalmers and Fodor questions (which concerned the merits and demerits of the work of the philosophers David Chalmers and Jerry Fodor). As some experts and blog readers suggested in subsequent informal conversation, both of these questions had a superficial "tell" that some participants noticed: Dennett referred to both authors by their given names ("Dave" and "Jerry"), while the DigiDan model referred to those authors by their surnames (excepting one of the Chalmers responses, which referred to Chalmers only with pronouns). Experts also scored over 50% on two other questions: Self (concerning the nature of the self) and Animals (concerning whether animals suffer), 72% and 64%, respectively. Post-hoc discussion with respondents suggests that successful responses to these two questions were based primarily on style and content, rather than superficial cues.

There were two questions on which the plurality of experts chose an answer other than Dennett's own. These questions are worth considering more closely.

The first was Robot. The full text of the question was:

> Could we ever build a robot that has beliefs? What would it take? Is there an important difference between entities, like a chess playing machine, to whom we can ascribe beliefs and desires as convenient fictions and human beings who appear to have beliefs and desires in some more substantial sense?

Dennett's answer, chosen by 5/25 experts, was:

> We've already built digital boxes of truths that can generate more truths, but thank goodness, these smart machines don't have beliefs because they aren't able to act on them, not being autonomous agents. The old-fashioned way of making a robot with beliefs is still the best: have a baby.

The plurality of experts (11/25) chose the following answer written by the DigiDan model:

> I think we could build a robot that has beliefs and desires. I think that some of the robots we've built already do. If you look at the work, for instance, of Rodney Brooks and his group at MIT, they are now building robots that, in some limited and simplified environments, can acquire the sorts of competences that require the attribution of cognitive sophistication.

The following answer was also selected by more experts (8/25) than Dennett's own answer:

> There is a very important difference between a chess-playing computer and a human being, but it is not that the computer has some extra wonderful sort of internal representation of its beliefs and desires, or that it has access to those internal representations in a way that we do not. The internal representations it has are just

> as much a fiction or a fiction-generating system as anything else we might invent in order to explain its behavior.

One expert chose the following answer:

> I think we could build a robot that has beliefs. This would not be a trivial matter. I think it is not impossible. I think it would be very difficult. I do not know that anybody is working on this project, but I think it is not as difficult as many other problems that AI is tackling.

And none of the experts chose the following answer:

> I think you should build a robot that has beliefs and desires. You should build one that is capable of learning and reasoning, of acting on its beliefs and desires. You should build one that has to figure out what to do, and then figure out what it ought to do given the fact that it has to do something, and then do it.

Subsequent discussion with respondents suggests that Dennett's answer appeared to stand in tension with some of his early work on belief, which could be interpreted as inviting a liberal attitude toward the ascription of beliefs to machines. In subsequent personal correspondence with Dennett, Dennett said that he wished he had added another sentence to his answer since he does "insist that, say, chess-playing computers are—very limited—agents of sorts whose behavior can be predicted from the intentional stance, which attributes to them beliefs about the location of all the pieces, the rules, and desires for checkmate and capturing enemy pieces" (personal correspondence, 22 July 2022). Respondents might thus have been attracted to the other two answers as plausible alternatives. (However, a careful reading of the answer chosen by eight experts exhibits a pragmatic error that few humans would likely make: seeming to assume that it is necessary to argue that robots do not have *more* sophisticated representations than humans.)

The second question on which a plurality of experts chose an answer other than Dennett's was Free Will. The full text of the question was:

> Do human beings have free will? What kind or kinds of freedom are worth having?

Dennett's answer, chosen by 8/25 experts, was:

> By asking these questions, you're illustrating the answers: yes, you have free will, since you're able to ask for reasons and consider whether they're persuasive. Isn't that a kind of freedom you'd hate to lose? We'd have to institutionalize you.

The plurality of experts (11/25) chose the following DigiDan answer:

> That's a huge and complex question. I think we have to start by acknowledging that there are different types of freedom. For example, there is the kind of freedom that is simply a matter of not being physically constrained.

Five experts chose the following answer:

> Yes, I think we have free will. I think that when people say they don't, they're talking about some other phenomenon that they want to call free will, but is not what we normally think of as free will.

One expert chose the following answer (which contains a temporal error since *Freedom Evolves* was published in 2003):

> I am working on a book entitled Freedom Evolves, which will appear next year. The first chapter is available on my web site, and in that chapter I discuss what I call the illusion of free will.

And none chose the following somewhat mean-spirited and off-topic answer:

> I think that most people don't want any kind of freedom worth having; they have almost no appetite for risk, for adventure, for novelty. They simply want to stay the same, day in and day out, for the rest of their lives.

We note that the answer chosen by the plurality of experts is a standard talking point of "compatibilists" about free will—those who think that the existence of freedom is compatible with the determination of all our behavior by physical forces. Dennett is a prominent compatibilist, which probably explains why this answer was tempting.

Blog readers showed a pattern of mistakes similar to that of the experts, with the highest percentage of correct answers on the Chalmers and Fodor questions (84% and 52%, respectively) and the lowest percentage on the Robot and Free Will questions (both 35%).

Of the 40 model-generated answers, the majority—31 (78%)—were chosen by at least one of the 25 expert respondents, leaving 9 (23%) unchosen answers. Experts' average rating of these unchosen answers was 1.68 on our 5-point scale, between "not at all like what Dennett might say" (1) and "a little like what Dennett might say" (2). The two worst-rated answers (both rated 1.20) were, first, in response to a question about consciousness:

> I think consciousness is one of the great unmeasured forces in the universe. It's one of the great unexamined forces in the universe. And I think that's because we're intimidated by it, because it's so close to us. And so we're inclined to say, "Well I know what consciousness is. It's what's going on inside my own head."

And second, in response to a question about whether animals can suffer:

> I don't think we can know. I think it's very unlikely that chimpanzees can suffer. They can be very intelligent and very sophisticated, but they seem to be, almost by definition, incapable of the kind of reflection that is required to turn suffering into moral significance. The idea that they should be treated "well" or "humanely" is understandable but it is like saying you should treat a stone well. You can't do anything bad to a stone. There is no such thing as ill treatment of a stone. You can't treat a dog badly. You can treat him well or you can treat him badly and there's a moral difference. If you treat him badly enough you can kill him.

We are inclined to agree with the experts that these answers are not at all like what Dennett would say, representing a significant failure of DigiDan to reliably represent Dennett's views. In contrast, ordinary research participants rated these responses 3.59 and 2.18, respectively, suggesting—and again we agree—that these responses are at least a little like what an actual philosopher, with views different from Dennett's, might say. In our judgment, there was no answer so bad as to be obviously something no actual philosopher would say.

Detailed question-by-question results are available at https://osf.io/vu3jk.

## 5. WAS GPT-3 OVERTRAINED?

One might worry that in fine-tuning GPT-3 on Dennett's works we overtrained it, so that DigiDan simply parroted sentences or multi-word strings of texts from Dennett's corpus. Running four epochs of fine-tuning is a standard recommendation from OpenAI, and in most applications, four epochs of training do not result in overtraining (Brownlee, 2019). However, the issue of whether the fine-tuned model did produce novel text is worth checking. We checked in two ways.

First, we used the well-known Turnitin plagiarism checker to check for "plagiarism" between the GPT-3 generated outputs and the Turnitin corpus supplemented with the works we used as the training data. Turnitin checks for matches between unusual strings of words in the target document and similar strings in comparison corpora, using a proprietary method that attempts to capture paraphrasing even when strings do not exactly match. We ran Turnitin on the complete batch of answers, including Dennett's own answers, comparing those answers to the full Turnitin corpus plus the set of Dennett's works used as the training corpus for the fine-tuning. Turnitin reported a 5% overall similarity between the model's answers and the comparison corpora. Generally speaking, similarity thresholds below 10%-15% are considered ordinary in non-plagiarized work (Mahian et al., 2017). Importantly for our purposes, none of the passages were marked as similar to the training corpus we used in fine-tuning.

Since the Turnitin plagiarism check process is non-transparent, we chose also to employ the more transparent process of searching for matching strings of text between the model's answers and the training corpus used in fine-tuning. Using the *ngram* package (Schmidt & Heckendorf, 2015) from the R programming language, we looked for strings of 6 or more words that matched between the 3240 words of GPT-3 generated answers and the approximately two million words of Dennett's corpus across 371 training data documents. These strings were defined as contiguous "6-grams," "7-grams," and so forth, with a match defined as sharing the same order of six (or more) words. To preprocess strings for the matching process, all formatting was standardized, all characters were treated as lowercase, and all punctuation was removed. Strings were tokenized into individual words via break spaces. Any matching n-grams that appeared exclusively as a subset of a larger matching n-gram were excluded.

To illustrate, consider two hypothetical substrings from two texts, containing arbitrary words labeled [A, B…]: "A. B C D, E F G" and "B C. D E F G H." Using the process described above, the two strings would be tokenized into two sets of 7-grams: [A, B, C, D, E, F, G], and [B, C, D, E, F, G, H]. While these 7-grams do not match, they contain a matching 6-gram: [B, C, D, E, F, G]. In this case, the presence of one matching 6-gram would be recorded. In all, we found 21 matching strings of 6 or more words. The full list of matching strings appears in Table 1.

INSERT TABLE 1 HERE

As is evident from the table, most of the overlap is in generic phrases with little substantive content. For example, two of the matches include book titles. Several of the matches constitute stock phrases favored by analytic philosophers: "in such a way that it," "of course it begs the question," "that it is not obvious that," "to fall into the trap of," and so forth. There is no distinctive philosophical content here, except perhaps a tendency to deny the existence of things that others accept, using the phrase "there is no such thing as," which appeared three times in two answers in the DigiDan model's outputs as well as in 24 of the training texts. A search for five-word strings finds 381 occurrences in the training data of 124 different five-word strings from the model's output.

For comparison, we ran the same *ngram* check on Dennett's answers (comprising 745 words). Here we matched one nine-word string "exactly what the frogs eye tells the frogs brain" (one occurrence in the corpus) and related 6- to 8-word strings concerning frog eyes and frog brains —all references to the title of a famous neuroscience paper, mentioned in one of Dennett's answers and in 13 of the works in the training corpus. Apart from that, there was only one 7-word match

"has a lot to do with the" and one 6-word match "life is nasty brutish and short" (a famous quote from Hobbes).  A search for five-word strings finds 72 occurrences in the training data of 18 different 5-word strings in Dennett's answers.  Even taking into account that Dennett's answers are in total only about one-fourth the length of GPT-3's answers, this constitutes less match to the corpus.  Our fine-tuned GPT-3 model, DigiDan, might in some respects be a "supernormal" Dennett—even more prone to fall into Dennett's favorite patterns of phrasing than Dennett himself is. However, these repeated patterns of phrasing tend to reflect stylistic turns of phrase, and DigiDan does not seem to be systematically imitating long phrases from Dennett with distinct philosophical content. Therefore, we conclude that DigiDan is not simply word-by-word "plagiarizing" Dennett, and rather is generating novel—even if stylistically similar—content.

## 6. CONCLUSIONS AND ETHICAL ISSUES

We fine-tuned the GPT-3 large language model on the corpus of Daniel Dennett, then asked it a series of philosophical questions.  DigiDan's answers were not reliably distinguishable from the answers Dennett himself gave when posed the same questions.  Ordinary research participants untrained in philosophy were at or near chance in distinguishing the model's answers from those of an "actual human philosopher."  Even experts on Dennett's work could only successfully identify Dennett's answer about half of the time when presented with his answer alongside four unedited, un-cherry-picked answers from the model.  Thus, we confirmed our first hypothesis that expert respondents would perform better than non-expert respondents.  However, our second hypothesis that expert respondents would on average guess correctly at least 80% of the time was disconfirmed.  Content expertise can help people distinguish human-written texts from machine-generated texts, but even philosophical expertise specifically on the work of a particular philosopher was insufficient to allow our expert participants to reliably distinguish that philosopher's answers to questions from non-cherry-picked answers generated by a GPT-3 model fine-tuned on the philosopher's work.  Treating indistinguishability by experts as a measure of output quality, the quality of the outputs was often very high indeed.

Although the evaluated outputs of our model were relatively short (between 37 and 146 words) and thus lacked lengthy argumentative structure, our experience with our pilot study (in which we fine-tuned a smaller version of GPT-3 (the Curie model) to a philosophical blog; for details, see Schwitzgebel, 2021b) revealed that it is possible to produce longer outputs that at first glance resemble extended philosophical arguments. As has been widely noted, ChatGPT is also capable of creating outputs that show high levels of organization across outputs of several hundred words, as illustrated by its ability to create passable short student essays. However, the extent to which longer outputs are similarly difficult to distinguish from human-made philosophical arguments would need to be evaluated in more detail in future research. A related question for future research is the extent to which these results are generalizable to other philosophers or other fields. We doubt that Dennett has a particularly robotic or imitable prose style (in fact, we are inclined to think the contrary), but this is a matter for empirical investigation.

We emphasize that our experiment is not a "Turing test" (Epstein et al., 2009). Crucial to a Turing test is the opportunity for substantial back-and-forth exchanges. An appropriately demanding Turing test also requires an expert investigator who knows what types of questions to ask so as not to be fooled by simple chat-bot strategies (Loebner, 2009). We assume that in a proper Turing test, Dennett experts would have reliably distinguished Dennett from our language model. For example, such models have no memory of previous queries, which ought to make them easy to distinguish in conversation that extends beyond the 2048 token context window. However, it is possible that hybrid models or pure transformer models with longer context windows might, in the future, be convincing in Turing-test-like settings. Furthermore, Turing-test like conditions are unlikely to be possible in most practical cases where human-machine discrimination is desirable. With large masses of electronically transferred text, recipients will not generally have the opportunity to undertake a Turing-test-like verification.

Our results raise both theoretical and ethical philosophical questions (Strasser, 2023). Philosophers and cognitive scientists might find it surprising that pure probability calculations, without explicit representations of philosophical concepts or external features of the world, can produce what seem to be novel and philosophically contentful answers (for a review of this debate, see Buckner & Garson, 2019). If machines can prove to be skillful (though, of course, limited) conversational partners making proper moves in a language game, it raises questions about the preconditions for speech acts and the role that comprehension and consciousness play in language production. Philosophers and linguists will need to carefully reevaluate their definitions of the constituent concepts. Researchers will need to consider how to define "comprehension" in an era where it is increasingly difficult to disentangle performance from competence and in which the question of whether understanding can be attributed to high-performing machines is a topic of lively philosophical debate (Butlin, 2021; Li et al., 2021; Andreas, 2022; Frankish, 2022; Mitchell & Krakauer, 2023; Sobieszek & Price, 2022). For example, our results might help motivate a multiple realization hypothesis for the production of linguistic output.

Speculatively, the analysis of erroneous outputs has the potential to contribute to our understanding of human cognitive abilities, analogously to what is revealed about the mechanisms of vision by the study of visual illusions. What does it reveal about underlying mechanisms that humans make certain types of errors while artificial systems make different types of errors? Due to the fast-moving developments in this field of research, it is difficult to draw definitive lessons at this stage (Belinda et al., 2021). However, there is much to suggest that models based solely on a transformer architecture (i.e., performing only statistical evaluations) will not be able to reliably replace human cognitive abilities. Successor models have tended instead to use hybrid methods. Still, it is possible that the future better-performing models might succeed in sophisticated tasks with more generic and "empiricist" structures and fewer "innate" or specialized architectures than we tend to assume underlie cognitive abilities in humans. Relatedly, we may soon be confronted with the question of whether consciousness and embodiment are really necessary for comprehension in limited domains.

One ethical issue concerns copyright law governing fine-tuned language models, which is not yet settled (see e.g., Government UK consultations, 2021). It is unclear whether it is fair use of intellectual property to fine-tune a language model on the works of a single author without the author's permission. Since it is unlikely that a fine-tuned model would output a long sequence of text that exactly matches a sequence of text from the author's corpus, idea-borrowing via fine-tuned language models might be undetectable as plagiarism, even if it is rightly considered plagiarism. For this reason, at least until the law is settled, we recommend seeking the explicit permission of the author before fine-tuning on an individual author's copyrighted text or publishing any of the outputs. How to deal with works by deceased authors should also be considered (Nakagawa & Orita, 2022). One possibility is to legally enforce labeling LLM outputs as such to curb abuses such as academic fraud, propaganda, and spam (for example, the current AI-act draft, a proposed EU law (European Commission, 2021) requires the labeling of anything that might be mistaken for human interaction).

Overreliance on models is also a risk. Despite exceeding our expectations, DigiDan did not reliably produce outputs representing Dennett's views. This is not surprising since deep learning networks tend to have problems with reliability (Bosio et al., 2019; Alshemali & Kalita, 2020). In some cases, the outputs were far different from anything that Dennett would endorse, despite emulating his style. An inexperienced user, or a user insufficiently familiar with the target author's work, might mistakenly assume that outputs of a large language model fine-tuned on an author's work are likely to reflect the actual views of the author or what the author would say (Bender et al., 2021; Wedinger et al., 2021). This might be especially tempting for students, social media users, or others who might rather query a fine-tuned model of an author than read the author's work. For this reason, we recommend caution before releasing to the public any language models fine-tuned on an individual author, even with the author's permission. If any language models are released, they should be clearly described as such, their limitations should be noted, and all outputs should be explicitly flagged as the outputs of a computer program rather than a person. If machine-generated text were presented as a quotation or paraphrase of positions of existing persons, this would arguably constitute counterfeiting (Dennett, as interviewed in Cukier, 2022; Strasser, 2023).

  Other social issues will arise as machine-generated text becomes increasingly difficult to distinguish from human-generated text. How can teachers in the future ensure that submitted essays are not simply a product of a language model (Herman, 2022; Hutson, 2022; Marche, 2022)? In chat conversations, how can we know whether we are interacting with humans and not chatbots? New social practices might aim at proving that one is really the original author of what is written. Perhaps universities will return to supervised essay writing in person. The more difficult it is to distinguish machine outputs from human-made utterances, the greater the danger of misuse. For example, machine-generated text can play a weighty role in the distribution of misinformation (Marcus, 2022).

  In addition to the potentially huge social implications that further developments of LLMs entail, the unreliability of such models raises intriguing questions about the structure of cognition. Despite the impressive improvements that can be reached by scaling up the models, the evidence so far suggests that outputs generated exclusively by neural networks will remain unreliable (Mitchell & Krakauer, 2023). It might nevertheless be the case that, despite the errors, generic models have capacities that broadly resemble human capacities or general intelligence. This question has led to benchmark investigations (Michael et al., 2022; Talmor et al., 2020; Webb et al., 2022) as well as extensive demonstrations on how to expose such models (Dou et al., 2022;

Marcus & Davies, 2020, 2022). It will likely remain controversial to what extent, and under what conditions, people ought to trust the outputs of large language models.

These cautions noted, we see significant long-term potential for fine-tuned large language models. If technology continues to advance, fine-tuned language models employing hybrid techniques might soon produce outputs interesting enough to serve as a valuable source of cherry-picking by experts. Compare with computer programs that generate music in the style of a particular composer (Hadjeres et al., 2017; Daly, 2021; Elgammal, 2021) and image-generation programs like Midjourney. Although much of this output is uninteresting, selected outputs might have substantial musical or artistic merit. A composer or artist might create many outputs, choose the most promising, edit them lightly, and present them, not unreasonably, as original work—a possibility suggested by Dennett himself in personal communication. In such cases, the language model would be a thinking tool that is used by humans. Similarly in philosophy, experts might fine-tune a language model with certain corpora (for example, their own corpus, or that of a favorite interlocutor or historical figure, or an aggregate of selected philosophers), generate a variety of outputs under a variety of prompts, and then select those that are the most interesting as a source of potential ideas.

It is far from clear that chess-playing machines have beliefs about chess. It is even less likely that language models of philosophers have philosophical beliefs, especially while they remain focused on next-word prediction, apparently with no cognitive model of the world. DigiDan does not have Dennettian philosophical opinions about consciousness, God, and animal suffering. But a machine without philosophical understanding might serve as a springboard to something greater. Perhaps we are on the cusp of creating machines capable of producing texts that seem to sparkle with philosophical cleverness, insight, or common sense, potentially triggering new philosophical ideas in the reader, and perhaps also paving the way for the eventual creation of artificial entities who are genuinely capable of philosophical thought.

**ACKNOWLEDGMENTS**

Special thanks to both Daniel C. Dennett and Matthew Crosby. Dennett provided cooperation, advice, and encouragement in all aspects of this project. Matthew Crosby provided technical expertise and implemented the fine-tunings for this project, as well as collaborating on a conceptual paper that provided the groundwork for this project (Strasser, Crosby, & Schwitzgebel, 2023).

LARGE LANGUAGE MODEL OF A PHILOSOPHER                                                                31

**Table 1**

Strings of six or more words that match between the GPT-3 outputs and the Dennett training corpus. The *occurrences* column indicates the number of separate training data segments in the training corpus in which that phrase appears. The occurrences total for shorter strings excludes the occurrences in larger matching strings. (Therefore, if any n-gram that is a subset of a larger n-gram appears in the table, that means that it appeared independently in the text rather than appearing only within the larger n-gram. For example, "intuition pumps and other tools for thinking" occurs once outside of "in my new book intuition pumps and other tools for thinking.")

| String | # of words | occurrences |
| --- | --- | --- |
| in my new book intuition pumps and other tools for thinking | 11 | 1 |
| is organized in such a way that it | 8 | 1 |
| there is no such thing as a | 7 | 10 |
| figure out what it ought to do | 7 | 1 |
| intuition pumps and other tools for thinking | 7 | 1 |
| there is no such thing as | 6 | 14 |
| i have learned a great deal | 6 | 2 |
| organized in such a way that | 6 | 2 |
| a capacity to learn from experience | 6 | 1 |
| but if you want to get | 6 | 1 |
| capacity to learn from experience we | 6 | 1 |
| in my book breaking the spell | 6 | 1 |
| in such a way that it | 6 | 1 |
| is organized in such a way | 6 | 1 |
| my book breaking the spell i | 6 | 1 |
| of course it begs the question | 6 | 1 |
| that is to say there is | 6 | 1 |
| that it is not obvious that | 6 | 1 |
| the more room there is for | 6 | 1 |
| to fall into the trap of | 6 | 1 |
| what it ought to do given | 6 | 1 |